\documentclass{article}
\usepackage{ijcai21}

\usepackage{times}

\usepackage{soul}
\usepackage{url}
\usepackage[hidelinks]{hyperref}
\usepackage[utf8]{inputenc}
\usepackage[small]{caption}
\usepackage{graphicx}
\usepackage{amsmath,amsthm}
\usepackage{amssymb}
\usepackage{booktabs}
\usepackage[svgnames,table]{xcolor}
\title{Learning Causal Models of Autonomous Agents using Interventions}

\author{
Pulkit Verma \And
Siddharth Srivastava\\
\affiliations
School of Computing, Informatics, and Decision Systems Engineering\\
Arizona State University, Tempe, AZ 85281, USA\\
\emails
\{verma.pulkit, siddharths\}@asu.edu
}

\usepackage{pict2e}
\usepackage{subcaption}
\usepackage{algorithm}
\usepackage{algorithmic}
\usepackage{tikz}
\usepackage{enumerate}
\usepackage[shortlabels]{enumitem}

\newcommand{\mc}[1]{\mathcal{#1}}
\newcommand{\citet}[1]{\citeauthor{#1}~[\citeyear{#1}]}
\newcommand{\h}{\mathcal{H}}
\newcommand{\ag}{\mathcal{A}}
\newcommand{\m}{\mathcal{M}}

\newcommand{\q}{\mathcal{Q}}

\newcommand{\mysssection}[1]{\noindent\textbf{#1}\hspace{10pt}}

\usepackage[cmtip,all]{xy}
\newcommand{\longsquiggly}{\xymatrix{{}\ar@{~>}[r]&{}}}

\newcommand{\actualcause}[5]{(\vec{#1}=\vec{#2}) \overset{(#3,\vec{#4})}{\longsquiggly} #5}

\theoremstyle{definition}
\newtheorem{theorem}{Theorem}

\newtheorem{definition}{Definition}
\newtheorem{lemma}{Lemma}
\newenvironment{sproof}{
	\proof}{\endproof}

\begin{document}

\maketitle

\begin{abstract}
One of the several obstacles in the widespread use of AI systems is the lack
of requirements of interpretability that can enable a layperson to ensure
the safe and reliable behavior of such systems.
We extend the analysis of an agent 
assessment module that lets
an AI system execute high-level instruction sequences in simulators and
answer the user queries about its execution of sequences of actions.
We show that such a primitive query-response capability is sufficient to
efficiently derive a user-interpretable \textit{causal} model of the system in
stationary, fully observable, and deterministic settings. We also 
introduce dynamic causal decision networks (DCDNs) that capture the causal 
structure of STRIPS-like domains. A comparative analysis of different classes of
queries is also presented in terms of the computational requirements needed to
answer them and the efforts required to evaluate their responses to learn the
correct model.
\end{abstract}

\section{Introduction}
The growing deployment of AI systems presents a pervasive problem of
ensuring the safety and reliability of these systems. The problem is
exacerbated because most of these AI systems are neither
designed by their users nor are their users skilled enough to understand
their internal working, i.e., the AI system is a
black-box for them. We also have systems that can
adapt to user preferences, thereby invalidating any design stage
knowledge of their internal model. Additionally, these systems have diverse
system designs and implementations. This makes it difficult to evaluate such
arbitrary AI systems using a common independent metric.

In recent work, we developed a non-intrusive system that allow for assessment
of arbitrary AI systems independent of their design and implementation. 
The Agent Assessment Module (AAM)~\cite{verma2021asking} is such a system
which uses active query answering to learn the action model of black-box autonomous
agents. It poses minimum requirements on the agent -- to have a rudimentary
query-response capability -- to learn its model using interventional queries.
This is needed because we do not intend these modules to hinder the development
of AI systems by imposing additional complex requirements or constraints on them.
This module learns the \textit{generalized dynamical causal model} of the agents 
capturing how the agent operates and interacts with its environment; and under what
conditions it executes certain actions and what happens after it executes them.

Causal models are needed to capture the behavior of AI systems as they help in
understanding the relationships among underlying causal mechanisms, and they also
make it easy to make predictions about the behavior of a system. E.g., consider a  
delivery agent which delivers crates from one location to another. If the agent has 
only encountered blue crates, an observational data-based learner might learn that 
the crate has to be blue for it to be delivered by the robot. On the other hand, a 
causal model will be able to identify that the crate color does not affect the
robot's ability to deliver it.

The causal model learned by AAM is user-interpretable as the model is learned in
the vocabulary that the user provides and understands. Such a module would also help
make the AI systems compliant with Level II assistive AI -- systems that make it
easy for operators to learn how to use them safely~\cite{srivastava2021unifying}.

This paper presents a formal analysis of the AAM, presents different types of query
classes, and analyzes the query process and the models learned by AAM. It also uses 
the theory of causal networks to show that we can define the causal properties
of the models learned by AAM -- in relational STRIPS-like
language~\cite{Fikes1971,McDermott_1998_PDDL,fox03_pddl}). We call this network
Dynamic Causal Decision Network (DCDN), and show that the models learned by AAM are 
causal owing to the interventional nature of the queries used by it.

\section{Background}

\subsection{Agent Assessment Module}
\begin{figure}
    \centering
    \includegraphics[width=\columnwidth]{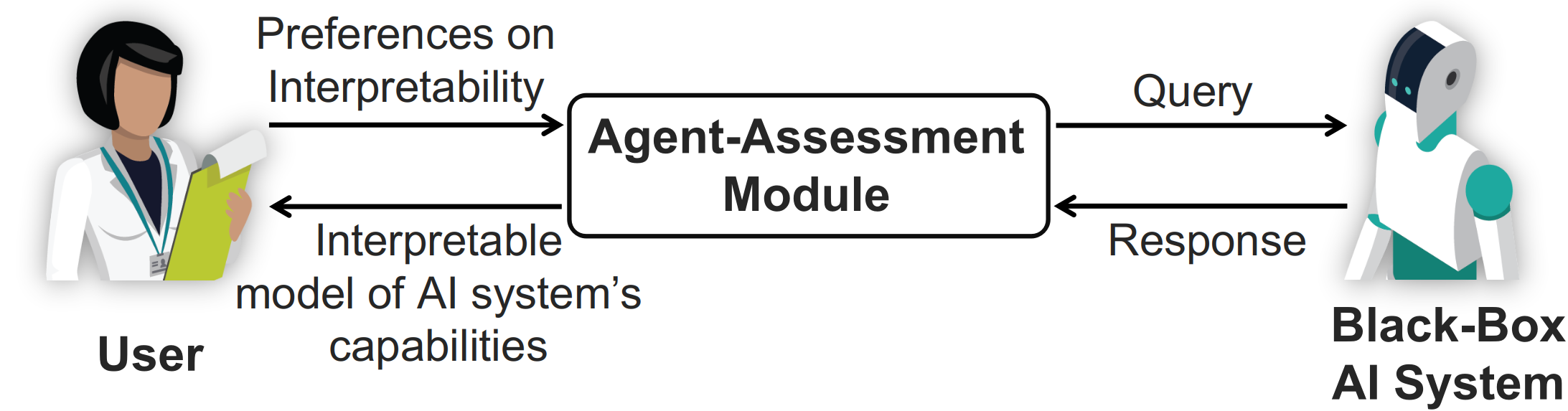}
    \caption{The agent-assessment module uses its user's
    preferred vocabulary, queries the AI system, and delivers a
    user-interpretable causal model of the AI system’s capabilities~\protect\cite{verma2021asking}}.
    \label{fig:aam}
\end{figure}

A high-level view of the agent assessment module is shown in Fig.\!~\ref{fig:aam}
where AAM connects the agent $\ag$ with a simulator and provides a sequence of
instructions, called a plan, as a \emph{query}. $\ag$ executes the plan in the
simulator and the assessment module uses the simulated outcome as the response to
the query. At the end of the querying process, AAM returns a user-interpretable
model of the agent. 

An advantage of this approach is that the AI system need not know the user
vocabulary or the modeling language and it can have any arbitrary internal  
implementation. Additionally, by using such a method, we can infer models of AI
systems that don't have such in-built capability to infer and/or communicate their 
model. Also, the user need not even know what questions are being asked as long as 
(s)he gets the correct model in terms of her/his vocabulary.
It is assumed that the user knows the names of the agent’s primitive actions. Even 
when they are not known, without loss of generality, the first step can be a
listing of the names of the agent’s actions.

Note that we can have modules like AAM with varying level of capabilities of
evaluating the query responses. This results in a trade-off between the evaluation
capabilities of the assessment modules and the computational requirements of the AI
systems to support such modules. E.g., if we have an assessment module with strong
evaluation capabilities, the AI systems can support them easily, whereas we might
have to put more burden on AI systems to support modules with weaker evaluation 
systems. To test and analyze this, we introduce a new class of queries in this
work, and study the more general properties of Agent Interrogation Algorithm (AIA) 
used by AAM. We also present a more insightful analysis of the complexity of the 
queries and the computational requirements on the agents to answer these queries.

\subsection{Causal Models}
In this work, we focus on the properties of the models learned by AIA, and show
that the models learned by AIA are causal. But prior to that, we must define what
it means for a model to be causal. Multiple attempts have been made to define causal
models~\cite{halpern2001causes,halpern2005causes,halpern_2015_modification}.
We use the definition of causal models based on \citet{halpern_2015_modification}.

\begin{definition}
\label{def:causal_models}
A \emph{causal model} $M$ is defined as a 4-tuple
$\langle \mc{U}, \mc{V}, \mc{R}, \mc{F} \rangle$
where $\mc{U}$ is a set of exogenous variables (representing factors outside 
the model's control), $\mc{V}$ is a set of endogenous variables (whose values are
directly or indirectly derived from the exogenous variables), $\mc{R}$  is a
function that associates with every variable $Y \in \mc{U} \cup \mc{V}$ a
nonempty set $\mc{R}(Y)$ of possible values for $Y$, and $\mc{F}$ is a function 
that associates with each endogenous variable $X \in \mc{V}$ a structural function
denoted as $F_X$ such that $F_X$ maps
$\times_{Z \in (\mc{U} \cup \mc{V} - \{X\})} \mc{R}(Z)$ to $\mc{R}(X)$.
\end{definition}

Note that the values of exogenous variables are not determined by the model, and a
setting $\vec{u}$ of values of exogenous variables is termed as a \emph{context} by
\citet{Halpern16_causality}. This helps in defining a causal setting as:

\begin{definition}
A \emph{causal setting} is a pair $(M,\vec{u})$ consisting of a causal model $M$ and context $\vec{u}$. 
\end{definition}

A causal formula $\varphi$ is true or false in a causal model, given a context.
Hence, $(M,\vec{u}) \models \varphi$ if the causal formula $\varphi$ is true in the 
causal setting $(M,\vec{u})$.

Every causal model $M$ can be associated with a directed graph, $G(M)$, in which 
each variable $X$ is represented as a vertex and the causal relationships between 
the variables are represented as directed edges between members of
$\mc{U} \cup \{\mc{V}\setminus X\}$ and $X$ \cite{pearl_2009_causality}. We use the
term causal networks when referring to these graphs to avoid confusion with the
notion of causal graphs used in the planning literature~\cite{Helmert04_Planning}.

To perform an analysis with interventions, we use the concept of \emph{do-calculus}
introduced in \citet{pearl_1995_do_calculus}. To perform interventions on a set of
variables $X \in \mc{V}$, do-calculus assigns values $\vec{x}$ to $\vec{X}$, and
evaluates the effect using the causal model $M$. This is termed as
\emph{do($\vec{X} = \vec{x}$)} action. To define this concept formally, we first
define \emph{submodels}~\cite{pearl_2009_causality}.

\begin{definition}
Let $M$ be a causal model, $X$ a set of variables in $\mc{V}$, and $\vec{x}$ a 
particular realization of $\vec{X}$. A \emph{submodel} $M_{\vec{x}}$ of $M$ is the 
causal model $M_{\vec{x}} = \langle \mc{U}, \mc{V}, \mc{R}, \mc{F}^{\vec{x}}
\rangle$ where $\mc{F}^{\vec{x}}$ is obtained from $\mc{F}$ by setting $X' = x'$
(for each $X' \in \vec{X}$) instead of the corresponding $F_{X'}$, and setting
$F^{\vec{x}}_Y = F_Y$ for each $Y \not\in X$.
\end{definition}

We now define what it means to intervene $\vec{X} = \vec{x}$ using the action
\textit{do}$(\vec{X} = \vec{x})$.

\begin{definition}
Let $M$ be a causal model, $X$ a set of variables in $V$, and $\vec{x}$ a particular
realization of $\vec{X}$. The effect of action \emph{do$(\vec{X} = \vec{x})$} on $M$
is given by the submodel $M_{\vec{x}}$.
\end{definition}

In general, there can be uncertainty about the effects of these interventions,
leading to probabilistic causal networks, but in this work we assume that 
interventions do not lead to uncertain effects.

The interventions described above assigns values to a set of variables, without 
affecting any another variable. Such interventions are termed as \textit{hard} 
(independent) interventions. It is not always possible to perform such interventions
and in some cases other variable(s) also change without affecting the causal 
structure~\cite{Korb04_varieties}. Such interventions are termed as \textit{soft} 
(dependent) interventions.

We can also derive the structure of causal networks using interventions in the real 
world, as interventions allow us to find if a variable $Y$ depends on another 
variable $X$. We use \citet{Halpern16_causality}'s definition of dependence and 
actual cause.

\begin{definition}
A variable $Y$ \emph{depends on} variable $X$ if there is some setting of all the 
variables in $\mc{U} \cup \mc{V} \setminus \{X,Y\}$ such that varying the value of
$X$ in that setting results in a variation in the value of $Y$.
\end{definition}

\begin{definition}
Given a signature $\mc{S} = (\mc{U}, \mc{V}, \mc{R})$, a \emph{primitive event} is
a formula of the form $X=x$, for $X \in \mc{V}$ and $x = \mc{R}(X)$. A \emph{causal
formula} is $[\vec{Y} \leftarrow \vec{y}]\varphi$, where $\varphi$ is a Boolean 
combination of primitive events, $\vec{Y} = \langle Y_1, Y_2,\dots Y_i\rangle$ are 
distinct variables in $\mc{V}$, and $y_i \in \mc{R}(Y_i)$. $\varphi$ holds if $Y_k$ 
would set to $y_k$, for $k= 1, \dots, i$.
\end{definition}

\begin{definition}
\label{def:causality}
Let $X \subseteq \mc{V}$ be the subset of exogenous variables $\mc{V}$, and $\varphi$ be a boolean causal formula expressible using variables in $\mc{V}$. 
$\vec{X}=\vec{x}$ is an \emph{actual cause} of $\varphi$ in the causal setting 
$(M,\vec{u})$, i.e., $\actualcause{X}{x}{M}{u}{\varphi}$, if the following conditions hold:

\begin{enumerate}[leftmargin=1cm,start=1,label={AC\arabic*.}]
\item $(M,\vec{u}) \models (\vec{X}=\vec{x})$ and $(M,\vec{u}) \models \varphi$.
\item There is a set $\vec{W}$ of variables in $\mc{V}$ and a setting $\vec{x}'$ of 
the variables in $\vec{X}$ such that if 
$(M,\vec{u}) \models \vec{W} = \vec{w}^*$, then
$(M,\vec{u}) \models [\vec{X}\leftarrow \vec{x}',\vec{W}\leftarrow 
\vec{w}^*]\neg\varphi$.
\item $\vec{X}$ is minimal; there is no strict subset $\vec{X}'$ of $\vec{X}$ such 
that $\vec{X}'=\vec{x}'$ satisfies conditions AC1 and AC2, where $\vec{x}'$ is the 
restriction of $\vec{x}$ to the variables in $\vec{X}$.
\end{enumerate}

AC1 mentions that unless both $\varphi$ and $\vec{X}=\vec{x}$ occur at the same 
time, $\varphi$ cannot be caused by $\vec{X}=\vec{x}$. 
AC2\footnote{\citet{Halpern16_causality} termed it as AC2(a$^m$)} mentions that 
there exists a $\vec{x}'$ such that if we change a subset $\vec{X}$ of variables 
from some initial value $\vec{x}$ to $\vec{x}'$, keeping the value of other 
variables $\vec{W}$ fixed to $\vec{w}^*$, $\varphi$ will also change. AC3 is a 
minimality condition which ensures that there are no spurious elements in $\vec{X}$.
\end{definition}

The following definition specifies soundness and completeness with respect to the actual causes entailed by a pair of causal models.


\begin{definition}
Let $\vec{\mc{U}}$ and 
$\vec{\mc{V}}$ be the vectors of exogenous and endogenous variables, respectively; and 
$\Phi$ be the set of all boolean causal formulas expressible over variables in $\mc{V}$.

A causal model $M_1$ is \emph{complete} with respect to another causal model $M_2$ 
if for all possible settings of exogenous variables, all the causal relationships 
that are implied by the model $M_1$  are a superset of the set of causal 
relationships implied by the model $M_2$, i.e., $\forall \vec{u} \in \vec{\mc{U}}, \forall \vec{X},\vec{X}'  \subseteq \vec{\mc{V}}, \forall \varphi, \varphi' \in \Phi, \exists \vec{x} \in \vec{X}, \exists \vec{x}' \in \vec{X}'$ s.t.
$\{ \langle \vec{X}, \vec{u}, \varphi, \vec{x}\rangle: \actualcause{X}{x}{M_2}{u}{\varphi} \} \subseteq
\{\langle \vec{X}', \vec{u}, \varphi', \vec{x}' \rangle: \actualcause{X'}{x'}{M_1}{u}{\varphi'} \}
$.

A causal model $M_1$ is \emph{sound} with respect to another causal model $M_2$ 
if for all possible settings of exogenous variables, all the causal relationships 
that are implied by the model $M_1$  are a subset of the set of causal 
relationships implied by the model $M_2$, i.e., $\forall \vec{u} \in \,\, \vec{\mc{U}}, \forall \vec{X},\vec{X}'  \subseteq \vec{\mc{V}}, \forall \varphi, \varphi' \in \Phi, \exists \vec{x} \in \vec{X}, \exists \vec{x}' \in \vec{X}'$ s.t.
$\{ \langle \vec{X}, \vec{u}, \varphi, \vec{x}\rangle: \actualcause{X}{x}{M_1}{u}{\varphi} \} \subseteq
\{\langle \vec{X}', \vec{u}, \varphi', \vec{x}' \rangle: \actualcause{X'}{x'}{M_2}{u}{\varphi'} \}
$.

\end{definition}

\subsection{Query Complexity}
In this paper, we provide an extended analysis of the complexity of the queries that
AIA uses to learn the agent's model. We use the complexity analysis of relational 
queries by \citeauthor{vardi82_complexity}~[\citeyear{vardi82_complexity},
\citeyear{vardi_95_on}] to find the membership classes for data, expression, and 
combined complexity of AIA's queries. 

\citet{vardi82_complexity} introduced three kinds of complexities for relational 
queries. In the notion of query complexity, a specific query is fixed in the 
language, then \textit{data complexity} -- given as function of size of databases --
is found by applying this query to arbitrary databases. In the second notion of 
query complexity, a specific database is fixed, then the \textit{expression 
complexity} -- given as function of length of expressions -- is found by studying 
the complexity of applying queries represented by arbitrary expressions in the 
language. Finally, \textit{combined complexity} -- given as a function of combined 
size of the expressions and the database -- is found by applying arbitrary queries 
in the language to arbitrary databases.

These notions can be defined formally as follows \citet{vardi_95_on}: 

\begin{definition}
The complexity of a query is measured as the complexity of deciding if $t \in Q(B)$,
where $t$ is a tuple, $Q$ is a query, and $B$ is a database.
\begin{itemize}
    \item The \emph{data complexity} of a language $\mc{L}$ is the complexity of the
    sets $Answer(Q_e)$ for queries $e$ in $\mc{L}$, where $Answer(Q_e)$ is the 
    answer set of a query $Q_e$ given as: \\
    $Answer(Q_e) = \{(t,B)\,|\, t \in Q_e(B)\}$.
    \item The \emph{expression complexity} of a language $\mc{L}$ is the complexity 
    of the sets $Answer_\mc{L}(B)$, where $Answer_\mc{L}(B)$ is the answer set of a 
    database $B$ with respect to a language $\mc{L}$ given as: \\$Answer_\mc{L}(B) =
    \{(t,e)\,|\,e \in \mc{L} \text{ and } t \in Q_e(B)\}$.
    \item The \emph{combined complexity} of a language $\mc{L}$ is the complexity of
    the set $Answer_\mc{L}$, where $Answer_\mc{L}$ is the answer set of a language
    $\mc{L}$ given as: \\$Answer_\mc{L} = \{(t,B,e)\,|\,e \in \mc{L} \text{ and } t
    \in Q_e(B)\}$.
\end{itemize}
\end{definition}

\citeauthor{vardi82_complexity}~[\citeyear{vardi82_complexity},
\citeyear{vardi_95_on}] gave standard complexity classes for queries written in 
specific logical languages.
We show the membership of our queries in these classes based on the logical 
languages we write the queries in.

\section{Formal Framework}
\label{sec:framework}
The agent assessment module assumes that the user needs to estimate the agent's 
model as a STRIPS-like planning model represented as a pair
$\mathcal{M} = \langle \mathbb{P}, \mathbb{A} \rangle$, where
$\mathbb{P} = \{p_1^{k_1},\dots, p_n^{k_n} \}$ is a finite set of
predicates with arities $k_i$; $\mathbb{A} = \{a_1,\dots, a_k \}$
is a finite set of parameterized actions (operators). Each action
$a_j \in \mathbb{A}$ is represented as a tuple $\langle header(a_j),
pre(a_j), \emph{eff}(a_j) \rangle $, where $header(a_j)$ is the action
header consisting of action name and action parameters, $pre(a_j)$
represents the set of predicate atoms that must be true in a state
where $a_j$ can be applied, $\emph{eff}(a_j)$ is the set of positive
or negative predicate atoms that will change to true or false
respectively as a result of execution of the action $a_j$. Each
predicate can be instantiated using the parameters of an action, where
the number of parameters are bounded by the maximum arity of the
action. E.g., consider the action $\emph{load\_truck}(?v1, ?v2, ?v3)$
and predicate $at(?x, ?y)$ in the IPC Logistics domain. This predicate
can be instantiated using action parameters $?v1$, $?v2$, and $?v3$ as
$at(?v1, ?v1)$, $at(?v1, ?v2)$, $at(?v1, ?v3)$, $at(?v2, ?v2)$,
$at(?v2, ?v1)$, $at(?v2, ?v3)$, $at(?v3, ?v3)$, $at(?v3, ?v1)$,
and $at(?v3, ?v2)$. We represent the set of all such possible
predicates instantiated with action parameters as $\mathbb{P}^*$.

AAM uses the following information as input. It receives its instruction
set in the form of $header(a)$ for each $a \in \mathbb{A}$ from the agent.
AAM also receives a predicate vocabulary
$\mathbb{P}$ from the user with functional definitions of each predicate.
This gives AAM sufficient information to perform a dialog with $\ag$ about
the outcomes of hypothetical action sequences.

We define the overall problem of agent interrogation as follows. Given
a class of queries and an agent with an unknown model which can answer
these queries, determine the model of the agent. More precisely, an
\emph{agent interrogation task} is defined as a tuple $\langle
\mathcal{M}^\mathcal{A}, \mathbb{Q}, \mathbb{P}, \mathbb{A}_H\rangle$,
where $\mathcal{M}^\mathcal{A}$ is the true model (unknown to AAM) of
the agent $\ag$  being interrogated, $\mathbb{Q}$ is the class of
queries that can be posed to the agent by AAM, and $\mathbb{P}$ and
$\mathbb{A}_H$ are the sets of predicates and action headers that AAM
uses based on inputs from $\h$ and $\mathcal{A}$. The objective of
the agent interrogation task is to derive the agent model $\m^\ag$
using $\mathbb{P}$ and $\mathbb{A}_H$. Let $\Theta$ be the set of
possible answers to queries. Thus, strings $\theta^* \in \Theta^*$
denote the information received by AAM at any point in the query
process. Query policies for the agent interrogation task are functions
$\Theta^*\rightarrow \mathbb{Q}\cup \{\emph{Stop}\}$ that map sequences
of answers to the next query that the interrogator should ask. The
process stops with the \emph{Stop} query. In other words, for all
answers $\theta \in \Theta$, all valid query policies map all sequences
$x\theta$ to \emph{Stop} whenever $x\in \Theta^*$ is mapped to
\emph{Stop}. This policy is computed and executed online.

\mysssection{Running Example}
Consider we have a driving robot having a single action \emph{drive (?t ?s ?d)}, 
parameterized by the truck it drives, source location, and destination location. 
Assume that all the locations are connected, hence the robot can drive between any 
two locations. The predicates available are $at(?t\,\, ?loc)$, representing the 
location of a truck; and $src\_blue(?loc)$, representing the color of the source 
location. Instantiating $at$ and $src\_blue$ with parameters of the action 
\textit{drive} gives four instantiated predicates $at(?t\,\, ?s)$, $at(?t\,\, ?d)$, 
$src\_blue(?s)$, and $src\_blue(?d)$.

\section{Learning Causal Models}
The classic causal model framework used in Def.~\ref{def:causal_models} lacks the 
temporal elements and decision nodes needed to express the causality in planning 
domains.

To express actions, we use the decision nodes similar to Dynamic Decision
Networks~\cite{Kanazawa89_model}. To express the temporal behavior of planning 
models, we use the notion of Dynamic Causal Models~\cite{pearl_2009_causality} and 
Dynamic Causal Networks (DCNs)~\cite{Blondel17_identifiability}. These are similar 
to causal models and causal networks respectively, with the only difference that the
variables in these are time-indexed, allowing for analysis of temporal causal 
relations between the variables. We also introduce additional boolean variables to 
capture the executability of the actions. The resulting causal model is termed as a 
causal action model, and we express such models using a Dynamic Causal Decision 
Network (DCDN).

A general structure of a dynamic causal decision network is shown in
Fig.\!~\ref{fig:dcdn}. Here $s_t$ and $s_{t+1}$ are states at time $t$ and $t+1$ 
respectively, $a_t$ is a decision node representing the decision to execute action 
$a$ at time $t$, and executability variable $X^a_t$ represents if action $a$ is 
executable at time $t$. All the decision variables and the executability variables 
$X_t^a$, where $a\in \mathbb{A}$, in a domain are endogenous. Decision variables are
endogenous because we can perform interventions on them as needed.

\subsection{Types of Interventions}
To learn the causal action model corresponding to each domain, two kinds of 
interventions are needed. The first type of interventions, termed $\mc{I}_P$, 
correspond to searching for the initial state in AIA. AIA searches for the state 
where it can execute an action, hence if the state variables are completely 
independent of each other, these interventions are hard, whereas for the cases where
some of the variables are dependent the interventions are soft for those variables. 
Such interventions lead to learning the preconditions of an action correctly.

\begin{figure}
    \centering
    \includegraphics[width=0.5\columnwidth]{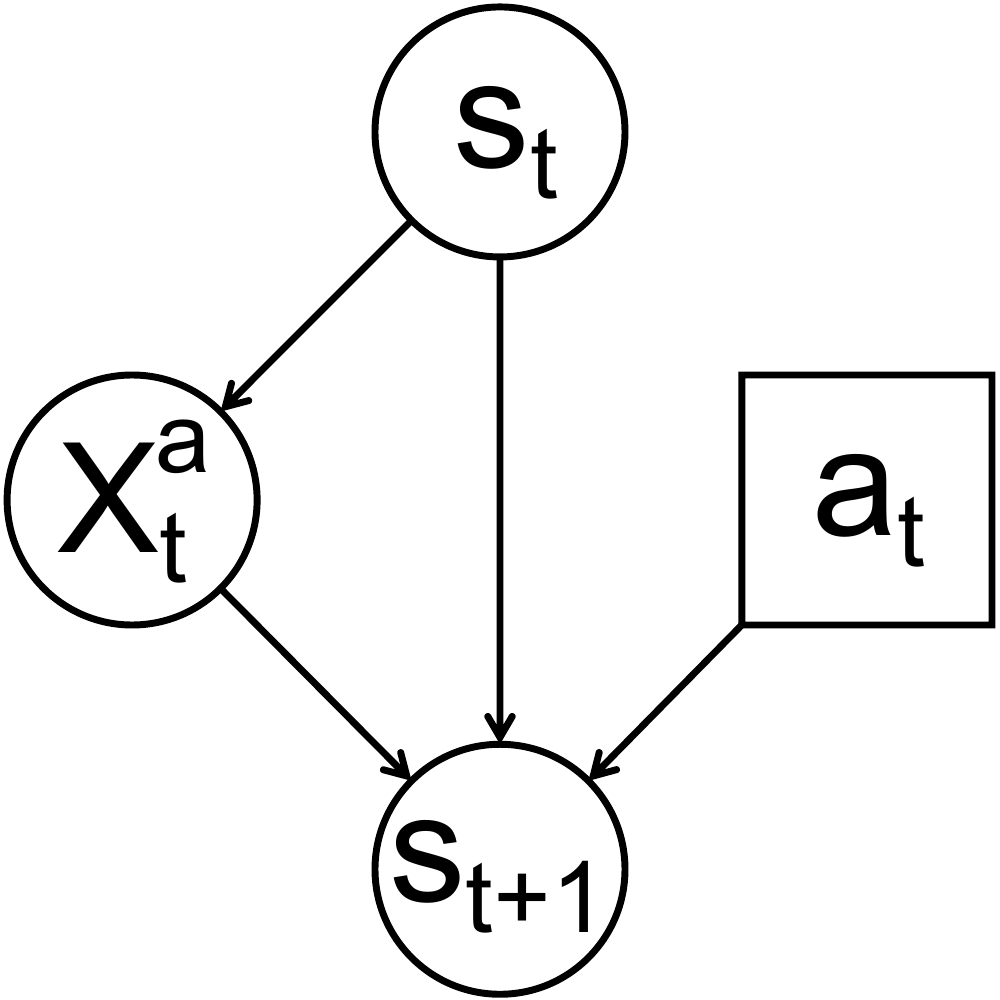}
    \caption{An example of a Dynamic Causal Decision Network (DCDN). $s_t$ and
    $s_{t+1}$ are states at time $t$ and $t+1$ respectively, $a_t$ is a decision
    node representing the decision to execute action $a$ at time $t$, and $X^a_t$
    represents if action $a$ is executable at time $t$.}
    \label{fig:dcdn}
\end{figure}

The second type of interventions, termed $\mc{I}_E$, are on the decision nodes, 
where the values of the decision variables are set to true according to the input 
plan. For each action $a_i$ in the plan $\pi$, the corresponding decision node with 
label $a_i$ is set to true. Of course, during the intervention process, the 
structure of the true DCDN is not known. Such interventions lead to learning the 
effects of an action accurately. As mentioned earlier, if an action $a$ is executed 
in a state $s_t$ which does not satisfy its preconditions, the variable $X^a_t$ will
be false at that time instant, and the resulting state $s_{t+1}$ will be same as
state $s_t$, signifying a failure to execute the action. Note that the state nodes
$s_t$ and $s_{t+1}$ in Fig.\!~\ref{fig:dcdn} are the combined representation of
multiple predicates.

We now show that the model(s) learned by AIA are causal models.

\begin{lemma}
Given an agent $\ag$ with a ground truth model $M^\ag$ (unknown to the agent interrogation algorithm AIA), the action model $M$ learned by AIA is a causal model consistent with Def.~\ref{def:causal_models}.
\end{lemma}
\begin{sproof}
We show a mapping between the components of the causal models used in 
Def.~\ref{def:causal_models} and the planning models described in 
Sec.\!~\ref{sec:framework}.
The exogenous variables $\mc{U}$ maps to the static predicates in the domain, i.e., 
the ones that do not appear in the effect of any action; $\mc{V}$ maps to the 
non-static predicates; $\mc{R}$ maps each predicate to $\top$ if the predicate is 
true in a state, or $\bot$ when the predicate is false in a state; $\mc{F}$ 
calculates the value of each variable depending on the other variables that cause 
it. This is captured by the values of state predicates and executability variables 
being changed due to other state variables and decision variables.

The causal relationships in the model $\m$ learned by AIA also satisfy the three 
conditions -- AC1, AC2, and AC3 -- mentioned in the definition for the actual cause 
(Def.~\ref{def:causality}). By Thm.\!~1 in \citet{verma2021asking}, AIA returns 
correct models, i.e., $\m$ contains exactly the same palm tuples as $\m^\ag$. 

This also means that AC1 is satisfied due to correctness of $\mc{I}_P$ -- a 
predicate $p$ is a cause of $X^a$ only when $p$ is precondition of action $a$; and  
$\mc{I}_E$-- a predicate $p$ is a caused by $X^a$ and $a$ only when $p$ is an effect
of action $a$. AC2 is satisfied because if any precondition of an action is not
satisfied, it will not execute (defining the relationship ``state variables
$\rightarrow X^a$''); or if any action doesn't execute, it won't affect the
predicates in its effects (defining the relationship ``$X^a \rightarrow$ state
variables''). Finally, AC3 is satisfied, as neither spurious preconditions are
learned by AIA, nor incorrect effects are learned.
\end{sproof}
We now formally show that the causal model(s) learned by AIA is(are) sound and 
complete.

\begin{theorem}
Given an agent $\ag$ with a ground truth model $M^\ag$ (unknown to the agent interrogation algorithm AIA), the action model $M$ learned by AIA is sound and complete with respect to $M^\ag$. 
\end{theorem}
\begin{sproof}
We first show that $M$ is sound with respect to $M^\ag$. Assume that some $\vec{X} = \vec{x}$ is 
an actual cause of $\varphi$ according to $M$ in the setting $\vec{u}$, i.e., $\actualcause{X}{x}{M}{u}{\varphi}$. Now by Thm.\!~1 in \citet{verma2021asking}, $\m$ contains exactly the same palm tuples as $\m^\ag$. Hence any palm tuple that is present in $M$ will also be present in $M^\ag$, implying that under the same setting $\vec{u}$ according to $M^\ag$ $\vec{X} = \vec{x}$ is an actual cause of $\varphi$.

Now lets assume that some $\vec{X} = \vec{x}$ is 
an actual cause of $\varphi$ according to $M^\ag$ in the setting $\vec{u}$, i.e., $\actualcause{X}{x}{M^\ag}{u}{\varphi}$. Now by Thm.\!~1 in \citet{verma2021asking}, $\m$ contains exactly the same palm tuples as $\m^\ag$. Hence any palm tuple that is present in $M^ag$ will also be present in $M$, implying that under the same setting $\vec{u}$ according to $M$ $\vec{X} = \vec{x}$ is an actual cause of $\varphi$. Hence the action model $M$ learned by the agent interrogation algorithm are sound and complete with respect to the model $M^\ag$.
\end{sproof}

\subsection{Comparison with Observational Data based Learners}
We compare the properties of models learned by AIA with those of approaches that 
learn the models from observational data only. For the methods that learn models in 
STRIPS-like the learned models can be classified as causal, but it is not necessary that they are sound with respect to the ground truth model $M^\ag$ of the agent $\ag$.
E.g., in case of the robot driver discussed earlier, these methods can 
learn a model where the precondition of the action drive is $src\_blue$ if all the 
observation traces that are provided to it as input had $src\_blue$ as true. This 
can happen if all the source locations are painted blue.
To avoid such cases, some of these methods run a pre-processing or a post-processing
step that removes all static predicates from the preconditions. However, if there is
a paint action in the domain that changes the color of all source locations, then
these ad-hoc solutions will not be able to handle that. Hence, these techniques may end 
up learning spurious preconditions as they do not have a way to distinguish between 
correlation and causations. 

On the other hand, it is also not necessary that the models learned by approaches using only observational data are complete with respect to the ground truth model $M^\ag$ of the agent $\ag$. This is because they may miss to capture some causal relationships if the observations do not 
include all the possible transitions, or contains only the successful actions. E.g.,
if we have additional predicates $city\_from(?loc)$, and $city\_to(?loc)$ in the 
domain, and all the observed transitions are for the transitions within same city, 
then the model will not be able to learn if the source city and destination city 
have to be same for driving a truck between them. 

Hence, the models learned using only observational data are not necessarily sound or complete, as they can learn causal relationships that are not part of set of actual causal relationships, and can also miss some of the causal relationships that are not part of set of actual causal relationships.
\citet{Pearl19_Seven} also 
points out that it is not possible to learn causal models from observational data 
only.

\subsection{Types of Queries}
\mysssection{Plan Outcome Queries}
\citet{verma2021asking} introduced plan outcome queries $\mathcal{Q}_{PO}$, 
which are parameterized by a state
$s_I$ and a plan $\pi$. Let $P$ be the set of predicates $\mathbb{P}^*$
instantiated with objects $O$ in an environment. $\mathcal{Q}_{PO}$
queries ask $\ag$ the length of the longest prefix of the plan $\pi$
that it can execute successfully when starting in the state $s_I
\subseteq P$ as well as the final state $s_F \subseteq P$ that this
execution leads to. E.g., ``Given that the truck $t1$ is at location $l1$, what
would happen if you executed the plan $\langle
drive(t1,l1,l2)$, $drive(t1,l2,l3)$, $drive(t1,l2,l1)
\rangle$?''

A response to such queries can be of the form ``I can execute the plan
till step $\ell$ and at the end of it truck $t1$ is at
location $l3$''. Formally, the response $\theta_{PO}$ for plan outcome
queries is a tuple $\langle \ell, s_{\ell} \rangle$, where $\ell$
is the number of steps for which the plan $\pi$ could be executed, and
$s_{\ell} \subseteq P$ is the final state after executing $\ell$
steps of the plan. If the plan $\pi$ cannot be executed fully according
to the agent model $\mathcal{M}^\mathcal{A}$ then $\ell < len(\pi)$,
otherwise $\ell = len(\pi)$. The final state $s_{\ell}\subseteq P$
is such that $\mathcal{M}^\ag \models \pi{[1:\ell]}(s_{I}) =
s_{\ell}$, i.e., starting from a state $s_{I}$, $\mc{M}^\ag$
successfully executed first $\ell$ steps of the plan $\pi$. Thus,
$\mathcal{Q}_{PO}: \mathcal{U} \rightarrow \mathbb{N} \times 2^P$,
where $\mc{U}$ is the set of all the models that can be generated using 
the predicates $P$ and actions $\mathbb{A}$, and $\mathbb{N}$ is the set
of natural numbers.

\vspace{0.1in}
\mysssection{Action Precondition Queries}
In this work, we introduce a new class of queries called \emph{action precondition
queries} $\mc{Q}_{AP}$. These queries, similar to plan outcome queries, are parameterized
by $s_I$ and $\pi$, but have a different response type.

A response to the action precondition queries can be either of the form ``I can execute the plan
completely and at the end of it, truck $t1$ is at
location $l1$'' when the plan is successfully executed, or of the form ``I can execute the plan
till step $\ell$ and the action $a_\ell$ failed because precondition $p_i$ was not satisfied'' when the plan is not fully executed. To make the responses consistent in all cases, we introduce a dummy action
$a_{\emph{fail}}$ whose precondition is never satisfied. Hence, the responses are always of the form,
``I can execute the plan till step $\ell$ and the action $a_\ell$ failed because precondition $p_F$ was not satisfied''. If $a_\ell$ is $a_{\emph{fail}}$ and $\ell = len(\pi)$, then we know that the original plan was
executed successfully by the agent. Formally, the response $\theta_{AP}$ for action precondition
queries is a tuple $\langle \ell, p_{F} \rangle$, where $\ell$
is the number of steps for which the plan $\pi$ could be executed, and
$p_F \subseteq P$ is the set of preconditions of the failed action $a_F$.  If the plan $\pi$ cannot be executed fully according
to the agent model $\mathcal{M}^\mathcal{A}$ then $\ell < len(\pi)-1$,
otherwise $\ell = len(\pi)-1$. Also, $\mathcal{Q}_{AC}: \mathcal{U} \rightarrow \mathbb{N} \times P$,
where $\mc{U}$ is the set of all the models that can be generated using 
the predicates $P$ and actions $\mathbb{A}$, and $\mathbb{N}$ is the set of natural numbers.

\section{Complexity Analysis}

Theoretically, the asymptotic complexity of AIA (with plan outcome queries) is $O(|\mathbb{P}^*| \times 
|\mathbb{A}|)$, but it does not take into account how much computation is needed
to answer the queries, or to evaluate their responses. This complexity just shows
the amount of computation needed in the worst case to derive the agent model by AIA.
Here, we present a more detailed analysis of the complexity of AIA's queries
using the results of relational query complexity by \citet{vardi82_complexity}.

To analyze $\q_{PO}$'s complexity, let us assume that the
agent has stored the possible transitions it can make (in propositional form)
using the relations $R (\emph{valid}, s, a, s', \emph{succ})$, where
$\emph{valid}, \emph{succ} \in \{\top, \bot\}$, $s, s' \in S$, $a \in A$;
and $N (\emph{valid}, n, n_+)$, where $\emph{valid} \in \{\top, \bot\}$, $n, n_+ \in
\mathbb{N}$, $0 \leq n \leq L$, and $0 \leq n_+ \leq L+1$, where $L$ is the maximum 
possible length of a plan in the $\q_{PO}$ queries. $L$ can be an arbitrarily large 
number, and it does not matter as long as it is finite. Here, $S$ and $A$ are sets 
of grounded states and actions respectively. $succ$ is $\top$ if the action was
executed successfully, and is $\bot$ if the action failed. $valid$ is $\top$ when
none of the previous actions had $succ=\bot$. This stops an action to change a state
if any of the previous actions failed, thereby preserving the state that resulted 
from a failed action. Whenever $succ=\bot$ or $valid=\bot$, $s=s'$ and $n=n_+$ 
signifying that applying an action where it is not applicable does not change the 
state.

Assuming the length of the query plan, $len(\pi)=D$, we can write a query in first order logic, equivalent to the plan outcome query as
\begin{align*}
    \{(s_D, &n_D) \,| \,\, \exists s_1, \dots, \exists s_{D-1}, \exists succ_1,
    \dots, \exists succ_{D-1}, \\
    &\exists n_1, \dots, \exists n_{D-1}
    R(\top,s_0, a_1, s_1, succ_1) \land\\ &R(succ_1, s_1, a_2, s_2, succ_2) \land
    \dots \land\\ &R(succ_{D-1}, s_{D-1}, a_D, s_D, \top) \land\\
    & N(\top, 0, n_1) \land N(succ_1, n_1, n_2) \land \dots \land\\ &N(succ_{D-1},
    n_{D-1}, n_D)\}
\end{align*}

The output of the query contains the free variables $s_D=s_\ell$ and $n_D=\ell$. 
Such first order (FO) queries have the expression complexity and the combined 
complexity in PSPACE~\cite{vardi82_complexity}.  The data complexity class of FO 
queries is $AC^0$~\cite{immerman1987expressibility}.

The following results use the analysis in \citet{vardi_95_on}.
The query analysis given above depends on how succinctly we can express the queries.
In the FO query shown above, we have a lot of spurious quantified variables. We can
reduce its complexity by using \textit{bounded-variable} queries.
Normally, queries in a language $\mc{L}$ assume an inifinite supply $x_1, x_2,\dots$
of individual variables. A \textit{bounded-variable} version $\mc{L}^k$ of the
language $\mc{L}$ is one which can be obtained by restricting the individual
variables to be among $x_1,\dots, x_k$, for $k>0$. Using this, we can reduce the
quantified variables in $FO$ query shown earlier, and rewrite it more succinctly as
an $FO^k$ query by storing temporary query outputs.

\begin{align*}
    E(succ,s,a,s',succ',n, n')\! = &R(succ,s,a,s',succ') \land\\& N(succ, n, n')\\
    \alpha_1(succ,s,a_1,s',succ',n, n')\! = &E(\top,s_0,a_1,s',succ',0,n')\\
\end{align*}
We then write subsequent queries corresponding to each step of the query plan as
\begin{align*}
    \alpha_{i+1}(&succ,s,a_{i+1},s',succ',n,n') = \\
    &\exists s_1, \exists succ_1, \exists n_1 \{E(succ,s,a_{i+1},s_1,succ_1,n_1) 
    \land\\
    &\hspace{0.2in} \exists s, \exists succ, \exists n [succ=succ_1 \land s=s_1 
    \land\\ 
    &\hspace{0.3in} n=n_1 \land 
    \alpha_{i}(succ,s,a_i,s',succ',n,n')]\}
\end{align*}
Here $i$ varies from $1$ to $D$, and the value of $k$ is 6 because of 6 quantified
variables -- $s, s_1, succ, succ_1, n,$ and $n_1$. This reduces the expression and 
combined complexity of these queries to ALOGTIME and PTIME respectively.
Note that these are the membership classes as it might be possible to write the 
queries more succinctly.

For a detailed analysis of $\q_{AP}$'s complexity, let us assume that the agent 
stores the possible transitions it can make (in propositional form) using the 
relations $R (\emph{valid}, s, a, s', \emph{succ})$, where $\emph{valid}, 
\emph{succ} \in \{\top, \bot\}$, $s, s' \in S$, $a \in A$; and $\mc{S} (p, s)$, 
where $p \in P$, $s \in S$. $\mc{S}$ contains $(p,s)$ if a grounded predicate $p$ is
in state $s$.

Now, we can write a query in first order logic, equivalent to the action
precondition query as:
\begin{align*}
    \{(p) \,| \,\, &(\forall s_1
    \,\mc{S}(p,s_1) \Rightarrow \exists s' R(\top, s_1, a_1, s', \top))\,\, \land\\
    &(\forall s_1\, \neg \mc{S}(p,s_1) \Rightarrow \forall s' R(\top, s_1, a_1, s', 
    \bot)) \}
\end{align*} 

This formulation is equivalent to the $FO^k$ queries with $k=2$. This means that the
data, expression and combined complexity of these queries are in complexity classes 
AC$^0$, ALOGTIME, and PTIME respectively.

The results for complexity classes of the queries presented above holds assuming 
that the agent stores all the transitions using a mechanism equivalent to
relational databases where it can search through states in linear time. For the
simulator agents that we generally encounter, this assumption almost never holds 
true. Even though both the queries have membership in the same complexity class, an
agent will have to spend more time in running the action precondition query owing
to the exhaustive search of all the states in all the cases, whereas for the plan 
outcome queries, the exhaustive search is not always needed.

Additionally, plan outcome queries place very little requirements on the agent to
answer the queries, whereas action precondition queries require an agent to use 
more computation to generate it's responses. Action precondition queries also force
an agent to know all the transitions beforehand. So if an agent does not know its 
model but has to execute an action in a state to learn the transition, action
precondition queries will perform poorly as agent will execute that action in all
possible states to answer the query. On the other hand, to answer plan outcome
queries in such cases, an agent will have to execute at most L actions (maximum
length of the plan) to answer a query.

Evaluating the responses of queries will be much easier for the action precondition queries, whereas evaluating the responses of plan outcome queries is not straightforward, as discussed in \citet{verma2021asking}.
As mentioned earlier, the agent interrogation algorithm that uses the plan outcome queries has asymptotic complexity $O(|\mathbb{P}^*| \times 
|\mathbb{A}|)$ for evaluating all agent responses. On the other hand, if an algorithm is implemented with action precondition queries, its asymptotic complexity for evaluating all agent responses will reduce to $O(|\mathbb{A}|)$. This is because AAM needs to ask two queries for each action. The first query in a state where it is guaranteed that the action will fail, this will lead AAM to learn the action's precondition. After that AAM can ask another query in a state where the action will not fail, and learn the action's effects. This will also lead to an overall less number of queries.

So there is a tradeoff between the computation efforts needed for evaluation of query responses and the computational burden on the agent to answer those queries.






\section{Empirical Evaluation}
\label{sec:experiments}

We implemented AIA with plan outcome queries in Python to evaluate the efficacy of our
approach.
In this
implementation, initial states were collected by making the agent perform random walks in
a simulated environment. We used a maximum of 60 such random initial
states for each domain in our experiments. 
The implementation is optimized to store the agent’s answers to queries;
hence the stored responses are used if a query is repeated.

We tested AIA on two types of agents: symbolic agents that use models from
the IPC (unknown to AIA), and simulator agents that report states
as images using PDDLGym~\cite{silver2020pddlgym}. All experiments were executed
on 5.0 GHz Intel i9-9900 CPUs with 64 GB RAM running Ubuntu~18.04.

The analysis presented below shows that AIA learns the correct model with
a reasonable number of queries, and compares our results with the closest
related work, FAMA~\cite{aineto2019learning}. We use the metric of
\textit{ model accuracy} in the following analysis: the number of
correctly learned palm tuples normalized with  the total number of palm
tuples in $\m^\ag$.

\subsection{Experiments with symbolic agents} We initialized the agent
with one of the 10 IPC domain models, and ran AIA on the resulting agent.
10 different problem instances were used to obtain average performance
estimates.

Table \ref{tab:domain_var} shows that the number of queries required
increases with the number of predicates and actions in the domain. We used
Fast Downward~\cite{Helmert06thefast} with LM-Cut
heuristic~\cite{Helmert2009LandmarksCP} to solve the planning problems.
Since our approach is planner-independent, we also tried using
FF~\cite{hoffmann2001} and the results were similar. The low variance shows
that the method is stable across multiple runs.

\begin{table}[t]
    
    \centering
    
    \rowcolors{2}{}{gray!13}
    \begin{tabular}{*6l}
        \toprule
        \textbf{Domain} & $\mathbf{|\mathbb{P}^*|}$ &$\mathbf{|\mathbb{A}|}$ & $\mathbf{|\hat{\q}|}$ & $\mathbf{t_\mu}$ \textbf{(ms)} & $\mathbf{t_\sigma}$ \textbf{($\mathbf{\mu}$s})\\
        \midrule
        Gripper      & 5   & 3  & 17  & 18.0 & 0.2 \\
        Blocksworld & 9   & 4  & 48  & 8.4 & 36\\
        Miconic     & 10  & 4  & 39  & 9.2 & 1.4\\
        Parking      & 18  & 4  & 63  & 16.5 & 806\\
        Logistics    & 18  & 6  & 68  & 24.4 & 1.73\\
        Satellite    & 17  & 5  & 41  & 11.6 & 0.87\\
        Termes       & 22  & 7  & 134 & 17.0 & 110.2\\
        Rovers      & 82   & 9  & 370  & 5.1 &  60.3 \\
        Barman       & 83   & 17  & 357  & 18.5  & 1605 \\
        Freecell     & 100 & 10 & 535 & 2.24$^\dagger$ & 33.4$^\dagger$ \\
        \bottomrule
    \end{tabular}
    \caption{The number of queries ($|\hat{\q}|$), average time per query ($t_\mu$), and variance of time per query ($t_\sigma$)  generated by AIA with FD. Average and variance are calculated for 10 runs of AIA, each on a separate problem.
    {\small $^\dagger$Time in sec. }}
    \label{tab:domain_var}
\end{table}

\subsubsection{Comparison with FAMA} We compare the performance of AIA with
that of FAMA in terms of stability of the models learned and the time taken
per query. Since the focus of our approach is on automatically generating
useful traces, we provided FAMA randomly generated traces of length 3 (the
length of the longest plans in AIA-generated queries) of the form used
throughout this paper ($\langle s_I,a_1,a_2,a_3,s_G\rangle$).

\begin{figure}[t]
    \centering
    \includegraphics[width=\columnwidth]{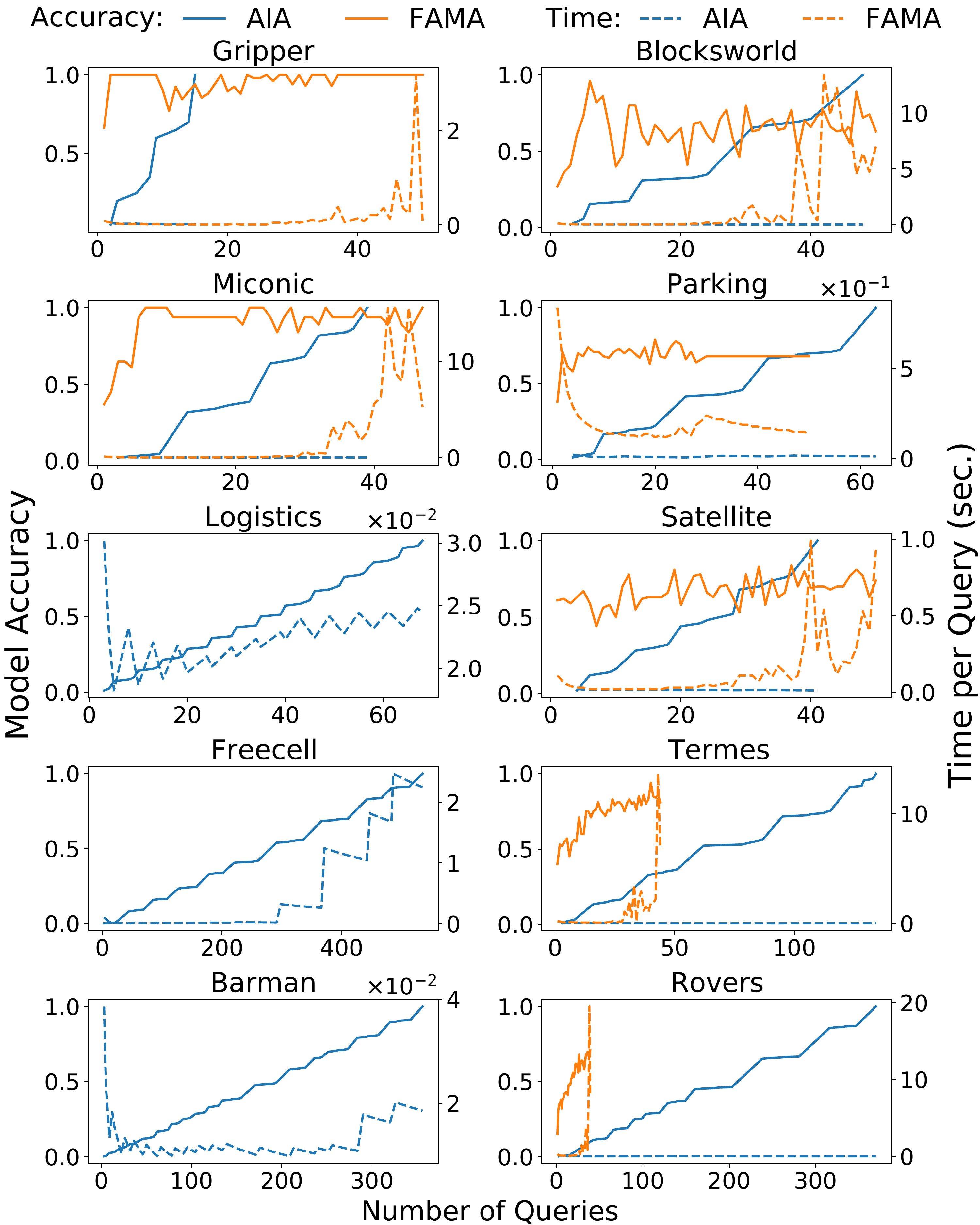}
    \caption{Performance comparison of AIA and FAMA in terms of model
    accuracy and time taken per query with an increasing number of queries.}
    \label{fig:graphs}
\end{figure}

Fig. \ref{fig:graphs} summarizes our findings. AIA takes lesser time per
query and shows better convergence to the correct model. FAMA sometimes
reaches nearly accurate models faster, but its accuracy continues to
oscillate, making it difficult to ascertain when the learning process
should be stopped (we increased the number of traces provided to FAMA until
it  ran out of memory). This is because the solution to FAMA's internal
planning problem introduces spurious palm tuples in its model if the input
traces do not capture the complete domain dynamics. For Logistics,
FAMA generated an incorrect planning problem, whereas for Freecell and
Barman it ran out of memory (AIA also took considerable time for Freecell).
Also, in domains with negative preconditions like Termes, FAMA was
unable to learn the correct model. We used
Madagascar~\cite{rintanen2014madagascar} with FAMA as
it is the preferred planner for it. We also tried FD and FF with FAMA, but
as the original authors noted, it could not scale and ran out of memory on
all but a few Blocksworld and Gripper problems where it was much slower
than with Madagascar.

\subsection{Experiments with simulator agents} AIA can also be used with
simulator agents that do not know about predicates and report states as
images. To test this, we wrote classifiers for detecting predicates from
images of simulator states in the PDDLGym
framework. This framework provides ground-truth PDDL models, thereby
simplifying the estimation of accuracy. We initialized the agent with one
of the two PDDLGym environments, Sokoban and Doors. AIA inferred the correct model in both cases and the
number of instantiated predicates, actions, and the average number of
queries (over 5 runs) used to predict the correct model for Sokoban were
35, 3, and 201, and that for Doors were 10, 2, and 252.

\section{Related Work}

One of the ways most current techniques learn the agent models is based on
passive or active observations of the agent's behavior, mostly in the
form of action traces~\cite{gil_94_learning,Yang2007,Cresswell09,Zhuo13action}.
\citet{jiminez_2012_review} and \citet{arora_2018_review} present comprehensive 
review of such approaches. FAMA~\cite{aineto2019learning} reduces model recognition to a planning
problem and can work with partial action sequences and/or state traces
as long as correct initial and goal states are provided. While
FAMA requires a post-processing step to update the learnt
model's preconditions to include the intersection of all states where
an action is applied, it is not clear that such a process would
necessarily converge to the correct model. Our experiments indicate
that such approaches exhibit oscillating behavior in terms of model
accuracy because some data traces can include spurious predicates, which
leads to spurious preconditions being added to the model's actions. 
As we mentioned earlier, such approaches do not feature interventions, and hence the models
learned by these techniques do not capture causal relationships correctly and
feature correlations.

\citet{Pearl19_Seven} introduce a 3-level causal hierarchy in terms of the classification of causal information
in terms of the type of questions each class can answer. He also notes that based on passive observations alone,
only associations can be learned, not the interventional or counterfactual causal relationships, regardless of the size of data.

The field of active learning~\cite{settles12} addresses the related
problem of selecting which data-labels to acquire for learning
single-step decision-making models using statistical measures of
information. However, the effective feature set here is the set of
all possible plans, which makes conventional methods for evaluating
the information gain of possible feature labelings infeasible.
In contrast, our approach uses a hierarchical abstraction
to select queries to ask, while inferring a multistep
decision-making (planning) model. Information-theoretic metrics could
also be used in our approach whenever such information is available.

\citet{Blondel17_identifiability} introduced Dynamical Causal Networks which extend the causal graphs to temporal domains, but they do not
feature decision variables, which we introduce in this paper.

\section{Conclusion} \label{sec:conclusion}


We introduced dynamic causal decision networks (DCDNs) to represent causal structures in
STRIPS-like domains; and showed that the models learned using the agent interrogation
algorithm are causal, and are sound and complete with respect to the corresponding unknown
ground truth models. We also presented an extended analysis of the
queries that can be asked to the agents to learn their model, and the
requirements and capabilities of the agents to answer those queries.

Extending the empirical analysis to action precondition queries, and extending our
predicate classifier to handle noisy state detection, similar
to prevalent approaches using classifiers to detect symbolic
states~\cite{Konidaris14,asai2017classical} are a few good directions for future
work. Some other promising extensions include replacing query and response
communication interfaces between the agent and AAM
with a natural language similar to \citet{Lindsay2017}, or learning other
representations like \citet{Zhuo14learnHTN}.



\section*{Acknowledgements}
This work was
supported in part by the NSF grants IIS 1844325, OIA 1936997, and ONR grant N00014-21-1-2045.



\bibliographystyle{named}
\bibliography{genplan21}

\end{document}